\documentclass[acmsmall]{acmart}

\AtBeginDocument{%
  }

\setcopyright{acmlicensed}
\copyrightyear{2025}
\acmYear{2025}
\acmDOI{XXXXXXX.XXXXXXX}

\acmJournal{JACM}
\acmVolume{1}
\acmNumber{2}
\acmArticle{3}
\acmMonth{5}

\usepackage{changes} 

\usepackage{algorithm}
\usepackage{algpseudocode}
\usepackage{amsmath}
\usepackage{booktabs}
\usepackage{multirow}
\usepackage{siunitx}
\usepackage{tikz}
\usepackage{paralist}
\usepackage{xspace}
\usepackage{url}
\usepackage{array}
\usepackage{adjustbox}
\usepackage{threeparttable}
\usepackage{graphicx}  
\usepackage{url}     
\usepackage{tabularx}

\usepackage{graphicx}
\usepackage{lineno}
\usepackage{subcaption}
\usepackage{titlesec}
\titlespacing*{\subsubsection}{0pt}{*1}{*1}  
\titleformat{\subsubsection}[block]{\normalfont\normalsize\bfseries}{\thesubsubsection}{1em}{} 

\newcommand{\modelname}{FuncGNN}
\acmSubmissionID{2024-P-2915}

\begin{document}

\title{FuncGNN: Learning Functional Semantics of Logic Circuits with Graph Neural Networks}


\author{Qiyun Zhao}
\email{xiaozhao_666@dlmu.edu.cn}






\begin{abstract}
 
As integrated circuit scale grows and design complexity rises, effective circuit representation helps support logic synthesis, formal verification, and other automated processes in electronic design automation. 
And-Inverter Graphs (AIGs), as a compact and canonical structure, are widely adopted for representing Boolean logic in these workflows.
However, the increasing complexity and integration density of modern circuits introduce structural heterogeneity and global logic information loss in AIGs, posing significant challenges to accurate circuit modeling.
To address these issues, we propose \modelname, which integrates hybrid feature aggregation to extract multi-granularity topological patterns, thereby mitigating structural heterogeneity and enhancing logic circuit representations.
\modelname\xspace further introduces gate-aware normalization that adapts to circuit-specific gate distributions, improving robustness to structural heterogeneity. 
Finally, \modelname\xspace employs multi-layer integration to merge intermediate features across layers, effectively synthesizing local and global semantic information for comprehensive logic representations.
Experimental results on two logic-level analysis tasks (i.e., signal probability prediction and truth-table distance prediction) demonstrate that \modelname\xspace outperforms existing state-of-the-art methods, achieving improvements of 2.06\% and 18.71\%, respectively, while reducing training time by approximately 50.6\% and GPU memory usage by about 32.8\%.

\end{abstract}


\maketitle

\section{Introduction}

As the scale of integrated circuits continues to expand and the complexity of designs keeps increasing, researching how to efficiently and accurately model and analyze circuits can help the field of Electronic Design Automation (EDA) achieve circuit design optimization, verification, and the optimization of automated processes~\cite{zhang2021circuit}. 
In the field of EDA, modeling complex logic circuits requires efficient and scalable representations, typically provided by Boolean networks.   
To address this need, AIGs simplify these networks into a compact, canonical form using two-input AND gates and inverters. 
By reducing the complexity of logic function processing, AIGs facilitate efficient logic synthesis, formal verification, and logic simulation, and thus serve as the standard intermediate representation in modern EDA workflows~\cite{mishchenko2006dag}.
This canonical structure supports scalable processing by standardizing logic representations across diverse EDA tasks. 
By capturing both structural and functional information, AIGs enable advanced applications, including circuit optimization, equivalence checking, and logic representation learning in deep learning-based EDA frameworks~\cite{huang2021machine,rapp2021mlcad,xu2025simtam}.

Conventional AIG processing methods, including structural hashing~\cite{subramanyan2013reverse}, functional propagation, and functionally reduced AIGs (FRAIGs)~\cite{mishchenko2005fraigs}, have enabled efficient logic synthesis and formal verification in EDA. 
However, these heuristic-based methods often struggle to capture the structural and functional complexities of large-scale logic circuits, leading to scalability and efficiency bottlenecks. 
This growing gap between the capabilities of traditional methods and the demands of modern circuit designs underscores the need for innovative AIG representation learning methods~\cite{fang2025survey,rapp2021mlcad}.

To address these limitations, recent advances in deep learning, particularly Graph Neural Networks~\cite{scarselli2008graph} (GNNs), have opened new avenues for circuit representation learning. 
GNNs excel at adaptively capturing complex dependencies between nodes in circuit graphs, enabling low-dimensional vector representations of logic gates. 
These representations have shown remarkable performance in various EDA tasks, including congestion prediction, power estimation, and testability analysis. 
In the context of AIG representation learning, GNN-based methods have gained traction by transforming circuit netlists into logic representations. 
For instance, DeepGate \cite{li2022deepgate} uses AIG-based representation with signal probability supervision for functional modeling. 
Similarly, GAMORA \cite{wu2023gamora} employs a multi-task framework to model circuit structure and logic function for gate-level recognition. 
Additionally, PolarGate \cite{liu2024polargate} introduces polarity embeddings and logic operators to enhance message passing. 
Furthermore, DeepGate3 \cite{shi2024deepgate3} combines GNNs and transformers with circuit-specific pretraining to enhance scalability.

Although existing methods have made notable progress, the growing integration density and functional complexity of modern integrated circuits have led to increasingly intelligent and large-scale logic designs. 
These trends introduce new demands on structural modeling and functional abstraction, giving rise to the following two challenges for existing methods:

\textbf{Challenge 1: Facing structural heterogeneity in AIGs.}
Due to the wide variation in circuit complexity and gate arrangements in real-world scenarios,
AIGs exhibit significant structural heterogeneity with diverse topologies and uneven gate distributions.  
Such heterogeneity complicates capturing consistent structural characteristics for AIG representation. 
Due to significant structural heterogeneity, existing methods struggle to capture consistent characteristics, making it difficult for existing GNN-based methods to generate stable and reliable  representations~\cite{yehudai2021local,joshi2022learning,bevilacqua2021size}.
Thus, achieving effective AIG representation to overcome structural heterogeneity remains a critical challenge.

\textbf{Challenge 2: Lacking efficient capture of global structural information in AIGs.}
With the increasing integration density of modern Field Programmable Gate Arrays (FPGA) designs, accurate circuit analysis increasingly depends on understanding global logic interactions across the entire AIG~\cite{alon2020bottleneck,xu2025novel}.
However, existing methods primarily rely on local message-passing mechanisms, which limits their ability to model long-range logic dependencies across large-scale AIGs. 
This often results in the loss of crucial global logic context, leading to reduced accuracy and limited scalability in downstream EDA tasks. 
While some methods introduce attention mechanisms to enhance global context modeling, they often suffer from significant computational overhead, making them inefficient for large-scale circuit analysis. 
Therefore, developing efficient and scalable representation methods that preserve and utilize global logic information remains a critical challenge.

To address these challenges, we propose \modelname, which combines local and global logic information to effectively address structural heterogeneity and global information loss, thereby enabling more efficient and accurate extraction of AIG logic semantics, which benefits EDA workflows such as design optimization and verification.
Specifically, \modelname\xspace comprises three key components: the Hybrid Feature Aggregation Component, the Global Context Normalization Component and the Multi-Layer Integration Component.

The \textbf{Hybrid Feature Aggregation Component} integrates diverse topological patterns, enhancing circuit property estimation by addressing AIG structural diversity. This approach improves logic representation accuracy, enabling robust analysis of heterogeneous circuit structures.
To further refine logic representations, the \textbf{Global Context Normalization Component} balances circuit-wide gate proportions, improving estimation accuracy. This mitigates uneven gate distributions, addressing structural heterogeneity (thus addressing Challenge 1).
Building on this, the \textbf{Multi-Layer Integration Component} synthesizes comprehensive logic relationships across the circuit, enhancing estimation accuracy. This captures global logic information, addressing information loss (thus addressing Challenge 2).

To effectively evaluate the model performance, we conducted experiments on a large-scale dataset of 9,933 AIG samples from four circuit benchmark suites to assess \modelname\relax’s performance in circuit property estimation. 
The results show \modelname\relax’s Signal Probability Prediction (SPP) accuracy and Truth-Table Distance Prediction (TTDP) improve by 2.06\% and 18.71\%, respectively, over the state-of-the-art methods. 
Additionally, \modelname\xspace enhances computational efficiency, improving logic representation for accurate circuit property estimation.

Overall, the main contributions of this work are threefold:

\begin{itemize}
    \item We introduce an innovative architecture tailored for functional representation learning on AIGs. This method leverages structural and logical properties of AIGs to capture functional dependencies effectively, addressing the over-squashing problem commonly encountered in deep AIG processing. By incorporating task-specific feature propagation, our method ensures robust learning of complex circuit behaviors, improving performance in tasks like SPP and TTDP.
    
    \item Extensive experiments demonstrate that \modelname\xspace achieves superior performance on heterogeneous AIGs, reducing loss by 2.06\% on the SPP task and 18.71\% on the TTDP task compared to state-of-the-art methods.
    Compared to the best models with equivalent parameter counts, \modelname\xspace maintains slightly higher accuracy  while reducing memory usage and runtime by 32.8\% and 50.6\%, effectively addressing learning difficulties in AIGs.
   
    \item We release the complete code of \modelname\xspace and the evaluation dataset to promote reproducibility and facilitate further investigation~\cite{funcgnn}.

\end{itemize}

\section{Related Work}
\label{sec:related}
As circuit scale and structural complexity continue to grow, circuit representation learning has become a key enabler in EDA. In this section, we categorize existing methods into two representative groups: attention-based approaches and message-passing-based approaches. We review their core designs, advantages, and limitations. This categorization clarifies the technical motivations underlying the design of \modelname.

\subsection{Attention-Based Methods}
 These methods capture long-range dependencies and structural patterns by leveraging attention mechanisms or structure-aware graph networks.
 DeepGate \cite{li2022deepgate} introduced AIG-based circuit representation with signal probability supervision, achieving functional-aware learning but with high computational cost.
 DeepGate2 \cite{shi2023deepgate2} improved upon this by incorporating truth-table supervision and a more expressive single-round GNN, which enhanced both modeling precision and training efficiency.
 DeepGate3 \cite{shi2024deepgate3} further extended the framework by combining GNNs and transformers with circuit-specific pretraining strategies, greatly improving scalability and generalization, albeit with increased resource requirements.
 HOGA \cite{deng2024less} proposed hop-aware aggregation and gated attention mechanisms, offering better scalability for large circuits, though it still struggled with modeling global contextual dependencies.

\subsection{Message-Passing Methods}
 Message-passing models emphasize local propagation and structural coherence through lightweight architectures.
 PolarGate \cite{liu2024polargate} introduced polarity embeddings and differentiable logic operators to enrich the message flow. While effective in certain tasks, it has difficulty generalizing across diverse AIG topologies.
 GAMORA \cite{wu2023gamora} proposed a multi-task framework that jointly models circuit structure and logic function, improving gate-level recognition. However, its capacity to represent global logic information remains limited.

 Despite their advances, current methods exhibit several limitations. Attention-based models often involve substantial computational overhead, limiting scalability to very large circuits. Message-passing approaches, though efficient, tend to lose global context and suffer from information degradation across deeper layers. Additionally, both categories face challenges in modeling the structural heterogeneity typical of AIG circuits. These issues reduce estimation accuracy and affect generalization across tasks.

 To address these challenges, \modelname\xspace introduces a novel architecture that integrates Hybrid Feature Aggregation, Global Context Normalization, and Multi-Layer Integration. These components are designed to maintain computational efficiency while improving functional accuracy and structural adaptability. Together, they enable robust and scalable AIG representation learning, offering support for downstream EDA tasks such as logic synthesis and equivalence checking.

\begin{figure*}[!t]
  \centering

  \includegraphics[width=1\linewidth]{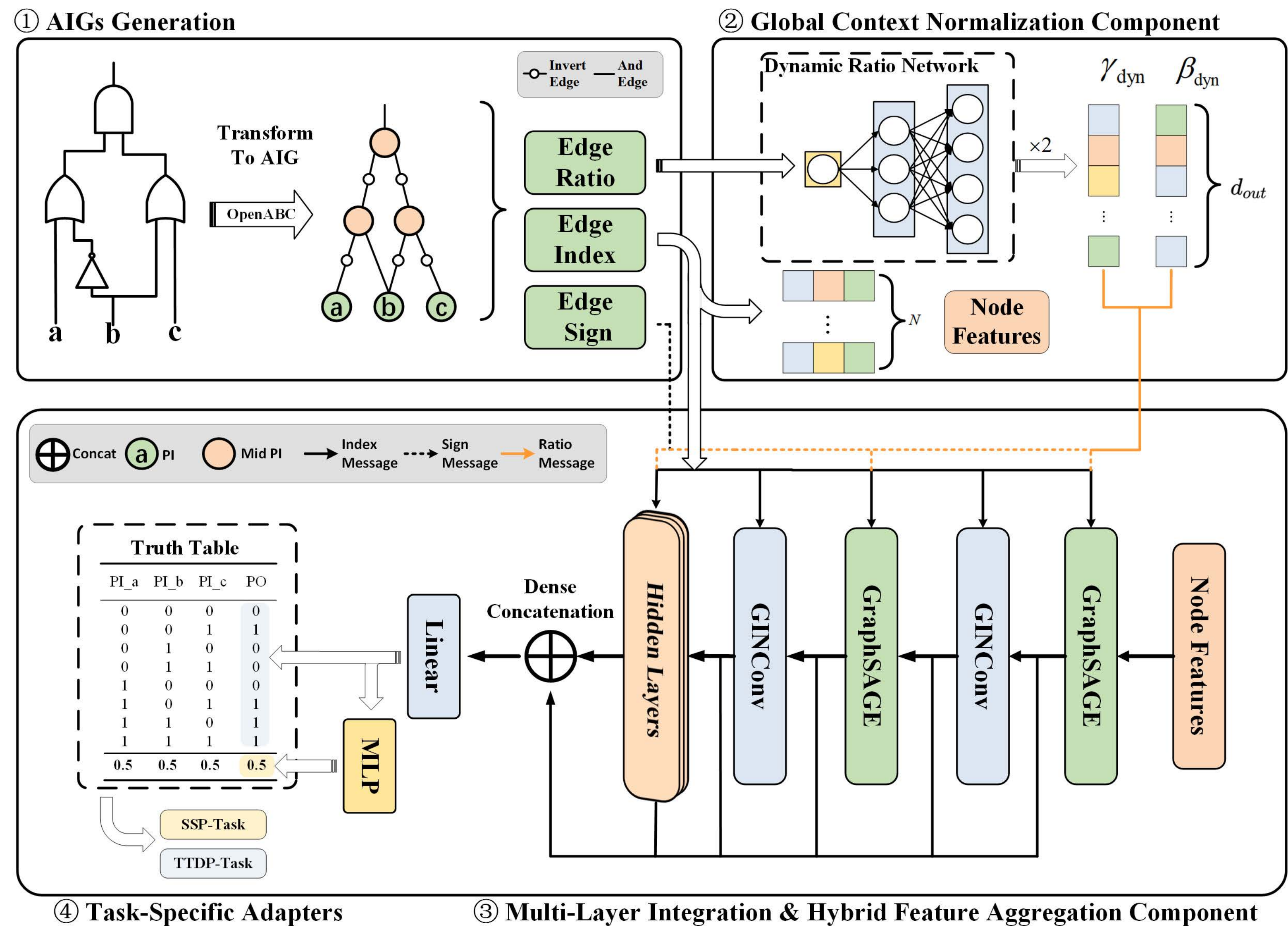}
\caption{The overall framework of \modelname}
\label{fig:3}
\end{figure*}

\section{\modelname\xspace Model}
\label{sec:method}


\subsection{Overview}\label{Overview}

The overall framework of \modelname\xspace is illustrated in Figure~\ref{fig:3}. \modelname\xspace is composed of three main components: the Hybrid Feature Aggregation Component, the Global Context Normalization Component and the Multi-Layer Integration Component.

The workflow begins with converting circuits into a unified AIG format using the open-source tool ABC~\cite{brayton2010abc}, yielding a standardized Boolean logic representation. 
Structural and logic information is then extracted from the AIGs to construct the model input. 
\modelname\xspace operates on three primary inputs: a node feature matrix, an edge tensor encoding logical relations (AND as +1, NOT as –1), and a global prior capturing the circuit-specific ratio of AND-to-NOT edges.
This input design enhances robustness against structural heterogeneity in AIGs by capturing diverse circuit topologies and gate distributions.

The node features are initially projected into a shared embedding space, while the edge tensor guides signal propagation and logic-aware aggregation during message passing.

To address structural heterogeneity, the \textbf{Hybrid Feature Aggregation Component} integrates two complementary feature extraction mechanisms with different granularities, alternating between them to enhance local and global information capture.
This dual design captures both local circuit patterns and node-level logic differences across varying topologies, effectively mitigating instability caused by circuit-specific gate distributions.

Further, the \textbf{Global Context Normalization Component} introduces a gate-aware normalization mechanism 
that integrates the AND-to-NOT ratio as a global prior. 
This module stabilizes training across diverse AIGs by reducing variance in feature distributions, further improving adaptability to heterogeneous circuit structures.

To alleviate the loss of global structural information, the \textbf{Multi-Layer Integration Component} performs multi-granularity logic information concatenation, preserving intermediate outputs from all processing stages. 
These are fused through a fully connected layer enabling the integration of logic information at multi-granularity across different structural depths. 
Furthermore, since the concatenation operation is computationally lightweight, this method maintains high efficiency while enhancing sensitivity to global logic patterns.

By combining these components, \modelname\xspace enables accurate, robust, and scalable circuit representation learning,and demonstrates superior performance over state-of-the-art methods on logic-level analysis tasks such as SPP and TTDP.

\subsection{Hybrid Feature Aggregation Component}
\label{GraphSAGE}

This section presents the architecture and operational workflow of the Hybrid Feature Aggregation Component, detailing its design principles for addressing structural heterogeneity in AIGs and supporting robust logic representation.

\begin{figure}[]
  \includegraphics[width=0.6\linewidth]{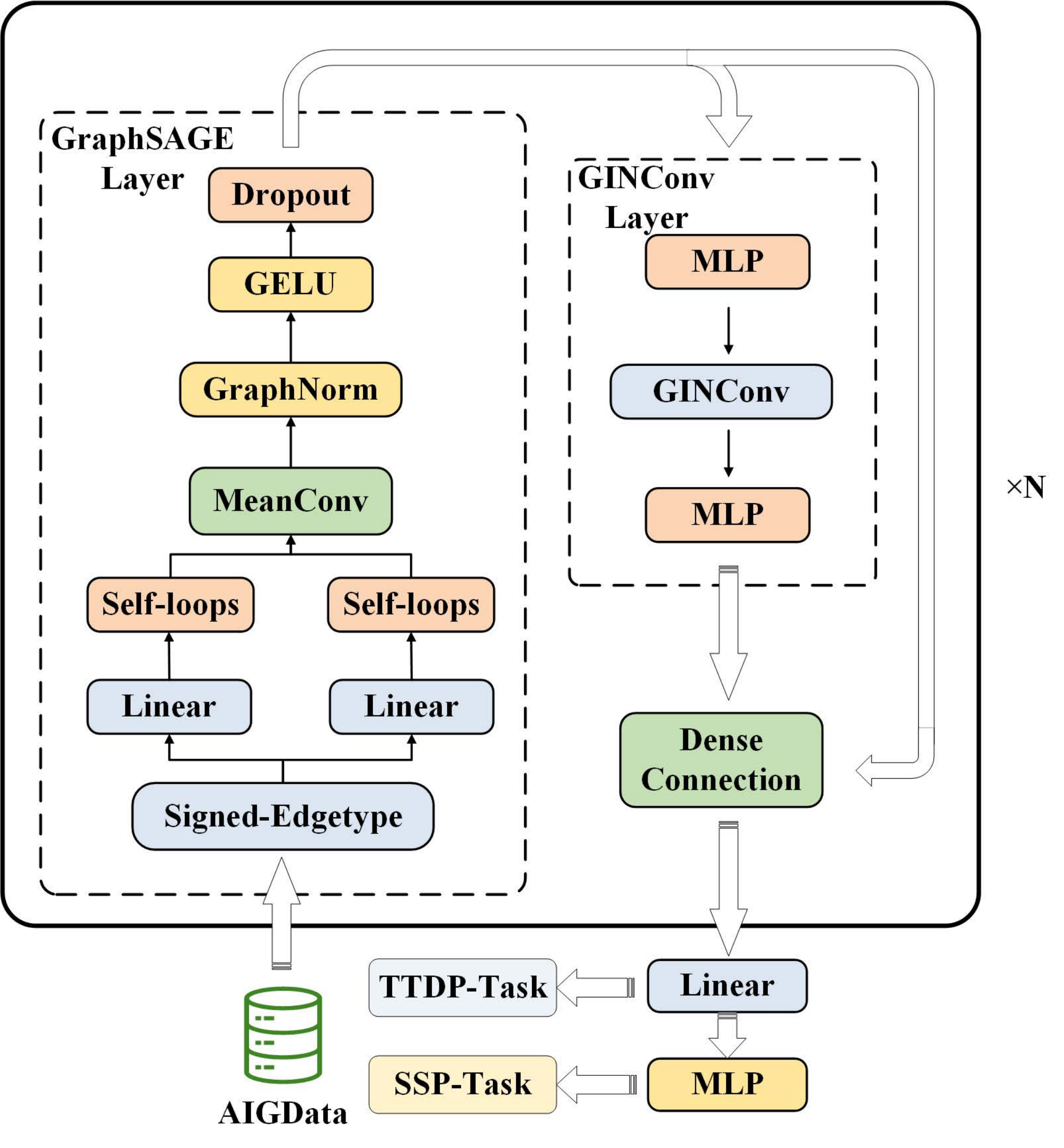}
\caption{The framework of Hybrid Feature Aggregation Component}
\label{fig:model_detail}
\end{figure}

AIGs exhibit significant structural heterogeneity, such as inconsistent AND-to-NOT gate ratios and diverse circuit topologies, which complicate the task of consistent feature extraction for logic synthesis and equivalence checking.
To address this, the Hybrid Feature Aggregation Component combines GraphSAGE~\cite{hamilton2017inductive}-based neighborhood aggregation with GINConv~\cite{xu2018powerful}-based nonlinear enhancement; its overall architecture is shown in Figure \ref{fig:model_detail}.

In particular, this component processes preprocessed graph data represented by a node feature matrix, an edge index matrix, and an edge sign vector. The edge sign vector distinguishes between AND and NOT relations, reflecting the logical structure of AIGs, and these elements are defined as follows:

\begin{equation}
X \in \mathbb{R}^{N \times d_{in}}, \quad E \in \mathbb{Z}^{2 \times E}, \quad \mathbf{s} \in \{-1, +1\}^{E}
\end{equation}
where $N$ denotes the number of nodes, $d_{in}$ denotes the dimensionality of the input features, $E$ denotes the number of edges; each column of the edge index matrix stores the source and target node indices, and each element of $s$ indicates the sign of the edge weight ($+1$ denotes an ‘AND’ relation, $-1$ denotes a ‘NOT’ relation).

Differentiated initial features are introduced to capture node-level logical differences, embedding circuit-specific attributes to distinguish logical roles across varying AND-to-NOT gate ratios. 
These features undergo a linear transformation to map into a unified hidden space, defined as:

\begin{equation}
X_{\text{trans}} = W X
\end{equation}
where $W$ denotes the transformed feature matrix, $d_{out}$ is the specified hidden‑space dimensionality, and is the trainable weight matrix. This equation represents a linear mapping of the input features from $d_{in}$ to $d_{out}$.

Then, a node retention mechanism incorporates self-loops to preserve each node’s logical identity during propagation, defined as:

\begin{align}
\mathbf{edge\_index} \sim \mathbf{edge\_index} \cup \{ (v, v) \mid v \in V \},\quad
\mathbf{edge\_sign} \sim \mathbf{edge\_sign} \cup \{ +1 \}
\end{align}
where $edge\_index$ denotes the edge‑index matrix and E the number of edges. 
\modelname\xspace add a self‑loop edge for every node $v \in V$ (i.e., $v$ connected to itself), introducing a new edge representing $v$’s connection to itself.

Signal propagation weights each neighbor’s contribution by its corresponding edge sign, distinguishing AND and NOT relations for accurate logical differentiation. For simplicity and efficiency, features for NOT edges are negated. 
This inter‑node message passing is thus defined as:
\begin{equation}
m_j(v) = \text{s}(e_{vj}) \cdot X_{\text{trans},j}, \quad \forall j \in N(v)
\end{equation}
where $m_j(v) $ is the message transmitted from neighbor $ j $ to node $ v$; $\sigma(v, j) \in \{–1, 1\} $is the edge‑sign weight for $(v, j)$; is $j$’s linearly transformed feature; and $N(v)$ is the set of $v$’s neighbors.

Feature smoothing integrates neighbor information, enhancing efficiency while maintaining robustness across heterogeneous AIGs. 
A node retention mechanism further integrates the node’s original features, mitigating information loss from over-smoothing, defined as:
\begin{equation}
h_v^{\text{agg}} = \dfrac{1}{|N(v)|} \sum_{j \in N(v)} m_j(v) + h_v^{\text{skip}}
\end{equation}
where $h_v^{\text{agg}}$ is the aggregated feature of node $v$, and $h_v^{\text{skip}}$ denotes the skip‑connection term using $v$’s original features.

Stabilization techniques, such as GELU~\cite{hendrycks2016gaussian} activation and dropout~\cite{srivastava2014dropout}, are employed to enhance numerical consistency and training robustness. GELU activation helps maintain stable feature distributions, while dropout mitigates overfitting by randomly deactivating neurons during training. By incorporating these techniques into the message aggregation framework, the overall node representation update can be formally expressed as follows:
\begin{equation}
h_{v}^{(k+1)} = \text{dropout} \left( \sigma \left( \text{Norm} \left( r, h_{v}^{(k)} + \sum\limits_{j \in \mathcal{N}(v)} \text{s}(e_{vj})\, h_{j}^{(k)} \right) \right) \right)
\end{equation}
where $h_v^{(k+1)}$ denotes the updated feature of node $v$ at layer $k+1$, and $h_v^{(k)}$ is its feature at the previous layer. $\text{s}(e_{vj})$ represents the semantic weight of the edge from node $j$ to node $v$, and $\mathcal{N}(v)$ is the set of neighbors of node $v$. $\text{Norm}$ refers to the enhanced conditional normalization operation, which will be discussed in the following section.

The term $\sigma$ refers to the GELU activation function, while $\text{dropout}$ indicates a mechanism that randomly discards features to prevent overfitting. 
The process involves aggregating neighbor signals, combining them with the node’s prior knowledge, and applying stabilization techniques. 
This yields stable and robust representations suitable for downstream processing across diverse circuit topologies.

To supplement stable feature extraction, a GINConv-based enhancement layer is attached after each aggregation stage, providing expressive logical enhancement to capture local circuit patterns and non-linear logical structures. 
A MLP is embedded within this enhancement, applying non-linear processing to each neighbor’s features. 
The MLP consists of two linear transformations interleaved with ReLU activations, introducing non-linearity and boosting expressiveness. The MLP kernel is defined as:

\begin{equation}
\text{MLP}(x) = W_4 \cdot \sigma(W_3 x + b_3) + b_4
\end{equation}
where $W_3$, $W_{4}$ are the trainable weight matrices of the two linear layers, $b_{3}$, $b_{4}$ are their bias vectors, $\sigma$ denotes the Rectified Linear Unit activation(ReLU), introducing nonlinearity and boosting expressiveness.

The enhancement operation aggregates these transformed features into a rich node representation, balancing the node’s own contribution. 
Combined with full-layer node retention, this design fuses multi-scale logical information across layers, heightening the capacity to represent complex logical structures inherent in AIGs. The GINConv operation is given by:

\begin{equation}
h_v^{(k+1)}
= (1 + \varepsilon)\,h_v^{(k)}\sum_{u \in \mathcal{N}(v)} h_u^{(k)}
\end{equation}
where $h_v^{(k+1)}$ is the updated feature of node $v$, $h_v^{(k)}$ is its feature from the previous GraphSAGE layer, $\varepsilon$ is a learnable scalar that balances the contribution of $v$’s own feature, $\sum_{u \in \mathcal{N}(v)} h_u^{(k)}$ sums the features of all neighbors of $v$.

By integrating GraphSAGE-based aggregation for stable feature extraction and GINConv-based enhancement for expressive local pattern capture, \modelname\xspace addresses varying AND-to-NOT ratios and diverse circuit topologies in AIGs. 
Therefore, the challenge of structural heterogeneity is alleviated to a certain extent. 
These features, however, exhibit cross-AIG variances, necessitating normalization to stabilize subsequent processing.

\subsection{Global Context Normalization Component}
This section elaborates on the architecture and operational workflow of the Global Context Normalization Component. Highlights the internal design of the component, which incorporates global logic statistics with the AND-to-NOT ratio. This design helps address inter-graph distribution shifts and stabilizes representation learning across heterogeneous AIGs.

To better address cross‑AIG variances and stabilize the training process, the Global Context Normalization Component leverages gate‑ratio information for graph normalization. 
Specifically, since the AND‑to‑NOT edge ratio exhibits significant disparities across different AIG instances, the model uses this proportion as prior knowledge. 
Incorporating this statistical data into the normalization stage reduces inter‑graph distribution shifts. 
Such conditional normalization mechanisms—akin to conditional BatchNorm (CBN)~\cite{de2017modulating} or Feature-wise Linear Modulation (FiLM)~\cite{perez2018film}—have been shown to enhance the use of global structural attributes and to capture more complex logical relationships.

In circuit benchmarks where all AIGs in the training dataset are generated uniformly (see Section~\ref{sec:dataset}), functional differences between logic circuits often appear as variations in their AND‑to‑NOT gate-ratio. 
Injecting these graph‑level functional attributes into the normalization step significantly improves functionality‑aware representation learning. 
From a circuit‑representation perspective, using AIG‑level global features as input further strengthens the ability to model intricate logic structures.

Specifically, enhanced GraphNorm adds two linear layers on top of standard GraphNorm~\cite{cai2021graphnorm}. 
These layers dynamically generate scaling and bias terms from the global ratio input, which are then added to the original weight and bias for adaptive normalization.
During normalization, each node feature is first standardized along the feature dimension as:

\begin{equation}
h = \dfrac{h - \mu}{\sqrt{\sigma^2 + \epsilon}}
\end{equation}
where $\mu $ and ${{\mathbf{\sigma }}^{2}}$ are the mean and variance of all node features in the current graph, and $\epsilon$ is a stability constant to prevent division by zero. 
This operation eliminates differences in mean and scale across node features.

Then, given the global ratio input $r$, we pass it through two dynamic ratio linear layers to produce a set of per‑feature scaling and bias parameters.

\begin{equation}
\gamma_{\text{dyn}} = \mathbf{W_1} \cdot \gamma \cdot r + \mathbf{b_1} \cdot \gamma ,\quad \beta_{\text{dyn}} = \mathbf{W_2} \cdot \beta \cdot r + \mathbf{b_2} \cdot \beta 
\end{equation}
where $W_i$ and $b_i$ are the learnable weight matrix and bias vector of those linear layers, used to generate dynamic scaling and bias coefficients.

These two dynamically generated terms are added to the original weight and bias, yielding the final scaling and bias as:

\begin{equation}
\gamma ={{\gamma }_{0}}+{{\gamma }_{\text{dyn}}},\quad \beta ={{\beta }_{0}}+{{\beta }_{\text{dyn}}}
\end{equation}
where ${\gamma }_{0}$ and ${\beta }_{0}$ are the globally learned scaling and bias parameters shared across all samples. ${\gamma }_{\text{dyn}}$ and ${\beta }_{\text{dyn}}$ are dynamically generated based on the current graph’s ratio $r$, allowing adaptation to each graph’s global characteristics. Their sum yields the final scaling $\gamma$ and bias $\beta$ applied to the normalized features.

Then, the features are conditionally normalized, producing the final output as:

\begin{equation}
{h_{\text{norm}}}=h\odot \gamma +\beta 
\end{equation}
where $\odot$ denotes element‑wise multiplication. By dynamically adjusting $\gamma$ and $\beta$ based on the AND‑NOT ratio, the Global Context Normalization Component becomes sensitive to cross‑AIG differences, thereby stabilizing training and enhancing the model’s representation of diverse logic circuits.

In summary, this component enables dynamic, circuit-aware normalization by incorporating global structural priors into the feature scaling process, thereby mitigating numerical and structural disparities. As a result, it significantly enhances the model’s stability and its capacity to generalize across diverse logic circuit topologies. 
The challenge of numerical and structural disparitie is effectively addressed.

\subsection{Multi-Layer Integration Component}\label{Dense}
This section introduces the Multi-Layer Integration Component, detailing its design and workflow to effectively capture and fuse multi-granularity logic information from all layers of the model. The discussion emphasizes how this component mitigates the over-squashing problem and preserves complementary features from shallow to deep layers, thus enhancing global representation learning in large-scale AIGs.

For large-scale AIGs, efficiently capturing global logic information while avoiding information degradation across layers is critical for accurate circuit representation. 
Different layers contribute complementary information at varying levels of granularity. 
Shallow aggregation layers focus on fine-grained neighborhood features, while deeper layers encode abstract global patterns. 
However, over-squashing—where representations from distant nodes become indistinguishable—can hinder the expressive capacity of deep models.

To address this, the Multi-Layer Integration Component preserves intermediate representations from all layers via full-layer dense concatenation, allowing the model to retain multi-granularity logic information across structural depths. 
These concatenated features are then fused through a learnable linear projection that adaptively emphasizes informative layers. 
This design not only mitigates global information loss caused by progressive over-squashing but also enhances sensitivity to hierarchical circuit patterns.

The architecture is designed without relying on attention-based mechanisms, instead using lightweight operations such as mean aggregation and linear transformation. This design ensures computational efficiency and scalability, enabling fast and robust extraction of full-graph semantics from complex and heterogeneous AIG structures.

Next, the concatenated feature vectors pass through a linear mapping layer that re‑weights and integrates multi‑scale information, reduces dimensionality, and preserves key signals. 
The resulting node features are then used in the TTDP loss computation. The process is defined as:

\begin{equation}
h_v^{\text{dense}} = W_{\text{fuse}} \cdot \text{concat} \left( h_v^{(1)}, h_v^{(2)}, \dots, h_v^{(L)} \right) + b_{\text{fuse}}
\end{equation}
where $h_v^{\text{dense}}$ denotes the node features after dense concatenation. The dense concatenation aggregates the output features from all $L$ layers—alternating between GraphSAGE and GINConv—into a higher-dimensional feature vector. $h_v^{(l)}$ refers to the output node features at the $l$-th layer, and $b$ denotes the bias term, which helps adjust the distribution of the fused features. The resulting output features are utilized for computations in the TTDP task or other similar downstream tasks.

Finally, these fused features enter an MLP‑based read‑out layer, which applies nonlinear transformations, LayerNorm, Dropout, and a Sigmoid activation to map outputs to the probability range (0,1),  yielding the SPP results.  The read-out MLP is formally defined as:

\begin{align}
   h_v^{(l+1)} = \text{dropout} \left( \sigma \left( \text{Norm} \left( W_l \cdot h_v^{(l)} + b_l \right) \right) \right), \quad \forall l \in \{0, \dots, L\}
\end{align}
where $h^{(l+1)}$ is the hidden feature at the $(l+1)$-th layer of the MLP, $\text{dropout}(\cdot)$ is applied to prevent overfitting by randomly deactivating a subset of neurons, and $\sigma(\cdot)$ introduces non-linearity. $\text{Norm}(\cdot)$ denotes the Layernorm, which stabilizes the training of SPP task, while $W_l$ and $b_l$ are the learnable weight and bias of the $l$-th layer. $L$ denotes the total number of layers in the MLP.

By retaining both stable local embeddings (from GraphSAGE) and fine‑grained non‑linear features (from GINConv), concatenating outputs from all layers, it combines early-stage local features with later-stage abstract representations, the dense connection layer captures comprehensive global structure. 
It leverages the complementary strengths of multi‑layer outputs and alleviates representation insufficiency in large, structurally diverse AIGs, while keeping the computational cost within an acceptable range. 
Integrating global context in this way significantly enhances the model’s learning and generalization capabilities for complex AIG graph prediction tasks. 
Therefore, the challenge of global structural deficiencies is effectively resolved.

In summary, the Multi-Layer Integration Component enables \modelname\xspace to combine stable local features with abstract global patterns by integrating outputs from every layer. This design alleviates global information loss caused by over-squashing, improves sensitivity to hierarchical circuit structures, and strengthens the model’s ability to generalize across structurally diverse and complex AIG graphs, all while maintaining computational efficiency.

\section{Experimental}
\label{sec:experiments}

This section outlines the experimental framework to evaluate \modelname\xspace on heterogeneous AIGs. Section~\ref{sec:rqs} formulates the research questions guiding the experiments. Section~\ref{sec:setup} details the dataset, state-of-the-art methods, and evaluation metrics, while Section~\ref{sec:train_details} elaborates on the experimental procedures and implementation specifics.

\subsection{Research Questions}\label{sec:rqs}

To systematically assess \modelname’s performance and robustness in capturing logical structures across diverse circuit topologies, the following research questions (RQs) are investigated:

    \textbf{RQ1}:Does \modelname\xspace improve SPP performance with lower resource consumption?

    \textbf{RQ2}: Does \modelname\xspace enhance accuracy on the TTDP task?

    \textbf{RQ3}: Does \modelname\xspace stabilize training for heterogeneous AIGs?

    \textbf{RQ4}: Does \modelname\xspace mitigate feature convergence in deep layers?

    \textbf{RQ5}: Are \modelname’s components necessary and effective?


RQ1 evaluates whether \modelname\xspace achieves higher accuracy in SPP with reduced model complexity, leveraging efficient feature integration. 
RQ2 assesses \modelname’s capability to capture fine-grained functional similarities in TTDP, as required for precise logic equivalence checking. 
RQ3 investigates whether \modelname\xspace stabilizes training across AIGs with diverse circuit topologies, addressing the challenge of structural heterogeneity. 
RQ4 examines whether \modelname\xspace overcomes feature convergence, ensuring robust capture of global logical structures in deep architectures. 
RQ5 validates the necessity and contribution of each component, such as hybrid aggregation and multi-layer integration, to the model’s effectiveness. 

\begin{figure}[!t]
  \centering
  \includegraphics[width=0.65\linewidth]{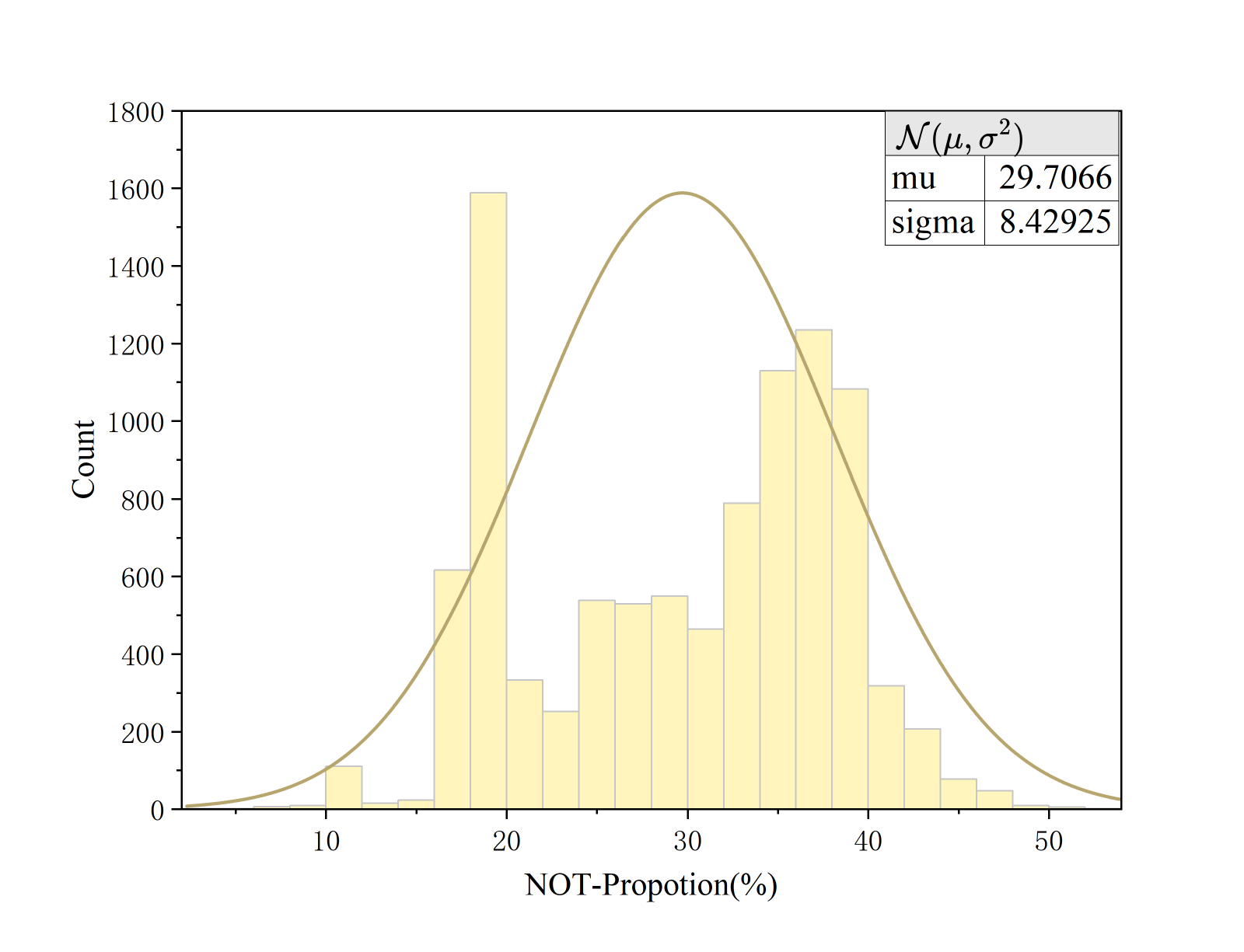}
  \caption{NOT Proportion Distribution Histogram}\label{fig:not_ratio}
\end{figure}

\begin{table}[]
\centering
\caption{The Statistics of AIG Datasets~\cite{li2022deepgate}}
\label{tab:dataset_stats}
\begin{tabular}{ccccc}
\toprule
Benchmark      & \#Subcircuits   & \#Node                  & \#Level    & \#Gate Ratio (A/N)         \\
\midrule
EPFL           & 828             & {[}52-341{]}            & {[}4-17{]}     &{[}1.27-30.4{]}      \\
ITC99          & 7,560           & {[}36-1,947{]}          & {[}3-23{]}      &{[}1.16-8.56{]}     \\
IWLS           & 1,281           & {[}41-2,268{]}          & {[}5-24{]}       &{[}0.90-22.2{]}    \\
Opencores      & 1,155           & {[}51-3,214{]}          & {[}4-18{]}       &{[}1.08-14.29{]}    \\
\midrule
\textbf{Total} & \textbf{10,824} & \textbf{{[}36-3,214{]}} & \textbf{{[}3-24{]}} &\textbf{{[}0.90-30.4{]}}\\
\bottomrule
\end{tabular}
\end{table}

\subsection{Experiment Settings}\label{sec:setup}

\subsubsection{Datasets}\label{sec:dataset}

For \modelname, the subcircuit dataset was constructed by extracting subcircuits from four benchmark suites—ITC'99~\cite{davidson1999characteristics}, IWLS'05~\cite{albrecht2005iwls}, EPFL~\cite{amaru2015epfl}, and OpenCore~\cite{opencores}—and converting them into a unified And-Inverter Graph (AIG) format using the ABC tool. Logic simulation with up to 100,000 random input vectors was performed to obtain accurate signal probabilities for each node. As presented in Table~\ref{tab:dataset_stats}, the dataset comprises 9,933 valid subcircuits, covering circuit sizes from tens to thousands of nodes with varying logic levels and exhibiting diverse circuit topologies and AND-to-NOT gate ratios. The column labeled "\#Subcircuits" specifies the number of subcircuits extracted from each benchmark. To visually illustrate the pronounced structural differences among AIGs in the dataset, a statistical analysis of the proportion of NOT nodes in the AIGs of all circuits was conducted. The histogram distribution is presented in Figure~\ref{fig:not_ratio}, revealing an irregular distribution of NOT gate proportions. Specifically, the mean NOT proportion is 29.7\%, with a standard deviation of 8.42. Furthermore, Table~\ref{tab:dataset_stats} highlights substantial variability in subcircuit scale (node counts ranging from 36 to 3,214), logic levels (3 to 24 levels), and subcircuit counts (828 to 7,560), underscoring the dataset’s high heterogeneity in topology, scale, and logical complexity. Notably, the "\#Gate Ratio" column shows a wide range from 0.90 to 30.4, reflecting large disparities in the relative proportions of AND and NOT gates across subcircuits. This multi-dimensional structural variability poses significant challenges to model robustness. Training and evaluating on such a structurally heterogeneous dataset highlights \modelname's strong generalization capabilities and validates its effectiveness in handling real-world AIG variability.

\subsubsection{State-of-the-Art Methods}\label{sec:baseline}
To evaluate the effectiveness of \modelname, it is compared with four state-of-the-art methods.

\textbf{DeepGate}~\cite{li2022deepgate} employs a circuit-specific attention mechanism that mimics logical computation processes and a backpropagation layer that accounts for logical effects, capturing both topological and functional properties of AIGs to achieve robust logical structure capture.

\textbf{DeepGate2}~\cite{shi2023deepgate2}, an enhanced version of DeepGate, introduces a function-aware learning framework. It leverages pairwise truth-table hardship as supplementary supervision and adopts an efficient single-pass GNN design to simultaneously capture structural and functional information of logic gates.

\textbf{DeepGate3}~\cite{shi2024deepgate3} builds on pretrained DeepGate2, utilizing a Refine Transformer to extract initial gate-level embeddings and further capture complex long-range dependencies among logic gates, excelling in handling large-scale circuit configurations.

\textbf{HOGA}~\cite{deng2024less} uses hop-wise feature aggregation with a gated self-attention module to integrate multi-hop features. This removes inter-node dependencies, simplifying training and enabling robust generalization across circuit topologies.

\textbf{PolarGate}~\cite{liu2024polargate} optimizes message passing mechanisms and model architecture to overcome functional representation bottlenecks, demonstrating notable improvements in training efficiency, prediction accuracy, and model flexibility compared to existing methods.

Additionally, classical graph neural network models \textbf{GCN}~\cite{kipf2016semi} and \textbf{GraphSAGE}~\cite{hamilton2017inductive} are included as comparative baselines for analysis. GCN~\cite{kipf2016semi} performs localized spectral convolutions that aggregate neighborhood features, while GraphSAGE~\cite{hamilton2017inductive} generates node embeddings by sampling and aggregating features from a node’s local neighborhood to enable inductive learning on large graphs.

\subsubsection{Evaluation Metrics}\label{sec:eval_metrics}

To assess the capability of \modelname\xspace in capturing logical structures within heterogeneous AIGs, two tasks are evaluated: SPP and TTDP. SPP, a fundamental task in AIG analysis~\cite{li2022deepgate}, measures the accuracy of predicting node signal probabilities across diverse circuit topologies. TTDP, introduced by DeepGate2~\cite{shi2023deepgate2}, evaluates the ability to capture fine-grained functional similarities by predicting the normalized Hamming distance between truth tables of node pairs. The performance is quantified using the following metrics, systematically evaluating prediction accuracy, computational efficiency, model complexity, and hardware overhead.

For the SPP task, the Mean Absolute Error (MAE) quantifies the average absolute difference between predicted and simulated signal probabilities for all nodes in the subcircuit dataset, as defined in Equation~\ref{eq:spp_mae}: \begin{equation}\label{eq:spp_mae}
Avg.Error_{SPP} = \frac{1}{N} \sum_{v \in \mathcal{V}} |y_v - \hat{y}_v| 
\end{equation} 
where $N$ is the total number of nodes, $V$ denotes the set of all nodes in the AIG, $y_v$ represents the simulated signal probability for node $v$, and $\hat{y}_v$ is the predicted signal probability.

For the TTDP task, the MAE measures the difference between the predicted and actual truth-table distances for sampled node pairs, capturing functional equivalence across AIGs with varying AND-to-NOT gate ratios. The computation involves calculating the normalized Hamming distance between truth tables and the distance between embedding vectors, followed by zero-normalization to align scales, as specified in Equations~\ref{eq:ttdp_start} to \ref{eq:ttdp_end}: 

\begin{equation}\label{eq:ttdp_start}
D_{(i,j)}^T = \frac{HammingDistance(T_i, T_j)}{length(T_i)}, \quad (i,j) \in \mathcal{N} \end{equation} 

\begin{equation} D_{(i,j)}^Z = 1 - CosineSimilarity(z_i, z_j) 
\end{equation}

\begin{equation}
\begin{aligned} D_{(i,j)}^{T^{\prime}} &= ZeroNorm(D_{(i,j)}^T), \quad D_{(i,j)}^{Z^{\prime}} = ZeroNorm(D_{(i,j)}^Z)  \end{aligned} \end{equation}
where $N$ is the set of sampled node pairs, $T_i$ and $T_j$ are the truth-table vectors for nodes $i$ and $j$, $HammingDistance(T_i, T_j)$ computes the number of differing bits between $T_i$ and $T_j$, and $length(T_i)$ is the length of the truth table. $z_i$ and $z_j$ denote the embedding vectors of nodes $i$ and $j$, $CosineSimilarity(z_i, z_j)$ measures their similarity, and $D_{(i,j)}^Z$ represents the embedding distance. $ ZeroNorm(\cdot)$ normalizes the distances to a zero-mean distribution, and $D_{(i,j)}^{T^{\prime}}$ and $D_{(i,j)}^{Z^{\prime}}$ are the normalized truth-table and embedding distances, respectively. 

Finally, we can define the MAE of TTDP task as:

\begin{equation}\label{eq:ttdp_end}
\text{Avg.Error}_{\text{TTDP}} = \dfrac{1}{N} \sum_{(i,j) \in \mathcal{N}} \left| D_{(i,j)}^{T^{\prime}} - D_{(i,j)}^{Z^{\prime}} \right|
\end{equation}
where $N$ is the number of node pairs, and $Avg.Error_{TTDP}$ is the average absolute difference between normalized distances.Where a lower MAE indicates higher accuracy and stronger capability in AIG representation learning.

Additional metrics include:
\begin{itemize} 

    \item \textbf{Parameter Count(Params)}: The total number of trainable parameters in \modelname, assessing model complexity and storage requirements. 
    
    \item \textbf{GPU Memory Usage(Mem)}: The maximum GPU memory usage in GB, where a lower value signifies less resource consumption, indicating the model’s feasibility in practical hardware environments.
    
    \item \textbf{Average Training Runtime (Avg.time)}: The average runtime per training epoch, where a shorter time reflects lower computational complexity on the same hardware,
    reflecting the computational efficiency and scalability of \modelname\xspace on large-scale AIG benchmarks. 
    
\end{itemize}

These four metrics—MAE, Avg.time, Parameter Count, and GPU Memory Usage—systematically evaluate \modelname’s performance from the perspectives of prediction accuracy, computational efficiency, model complexity, and hardware overhead, addressing the challenges of structural heterogeneity in large-scale AIGs.

\subsection{Training Details}\label{sec:train_details}

The experiments were conducted on a high-performance server equipped with an NVIDIA A800 GPU, running Ubuntu 22.04.5 LTS. For state-of-the-art methods, training hyperparameters were consistent with those reported in their respective papers, employing single-task training for up to 500 epochs. A default batch size of 128 was used, and the dataset was split into training, validation, and test sets with proportions of 0.05, 0.05, and 0.9, respectively. This extreme data split simulates real-world scenarios where labeled data is scarce, placing significant demands on a model’s generalization and learning ability.

For \modelname, the configuration of the Hybrid Feature Aggregation Component varies the number of layers $L$ across experiments, with a default of $L = 3$. The hidden dimension is set to 256, and the dropout rate is 0.1. In the Global Context Normalization Component, the output dimension of the mapped sum is set to 256. For the Multi-Layer Integration Component, the concatenated feature output dimension is 256, and the prediction MLP for the SPP task adopts a hidden dimension of 256 with a dropout rate of 0.1.

During training, initial feature vectors are generated as one-hot encodings based on node indices. The Adam optimizer is employed with an initial learning rate of 0.001 and a weight decay of $1 \times 10^{-4}$. An early stopping mechanism is applied, terminating training if no performance improvement is observed after 100 epochs, with the best-performing model selected as the final model.

\section{Results}\label{results}
\quad

In this section, five experiments were conducted to address the five research questions outlined in Section 4.

\subsection{Answer to RQ1: Performance on the SPP Task}
\quad
\textbf{Motivation.} The objective of this research question is to evaluate the performance of the proposed method on the SPP task and compare it with state-of-the-art methods.

\textbf{Methodology.} To assess the effectiveness of the method on the SPP task, all state-of-the-art methods were trained on the same dataset using default data splitting ratios and predefined hyperparameters. Performance comparisons were conducted based on four evaluation metrics: MAE, Params, Mem, and Avg.time. The layer count (L=n) denotes model complexity, with higher values indicating increased complexity and parameter counts. For DeepGate2, the two-stage training process described in its original paper was preserved.

\textbf{Results.} 
Table~\ref{tab:1} reports the experimental results on the AIG benchmark dataset. Overall, \modelname\xspace consistently outperforms all baselines in both predictive accuracy and resource efficiency, validating its design for scalable and functionality-aware AIG representation learning.

Compared to PolarGate (L=9), which uses a comparable parameter size, \modelname\xspace achieves a 2.06\% lower MAE, while also reducing memory consumption by 32.83\% and average training time by 50.65\%. This performance gain can be attributed to the Hybrid Feature Aggregation Component,
which replaces resource-intensive attention modules with lightweight aggregation, enabling more efficient learning of logic patterns without sacrificing accuracy.

When compared to DeepGate2, a strong attention-based baseline, \modelname\xspace shows even more significant improvements. It reduces the MAE by 57.01\%, with a dramatically smaller parameter count and lower memory usage. This is largely due to the introduction of the Global Context Normalization Component, which explicitly incorporates global logic statistics (e.g., AND-to-NOT ratio) into the learning process, allowing the model to generalize across structurally diverse circuits with fewer resources.

Against simpler models such as GCN and GraphSAGE, \modelname\xspace achieves 84.11\% and 88.94\% lower MAE respectively. These results highlight the benefits of \modelname’s Multi-Layer Integration Component, which retains and fuses multi-granularity information across layers. This allows the model to better capture both local structures and high-level circuit logic, addressing the over-squashing problem commonly seen in shallow message-passing networks.

In summary, \modelname\xspace not only achieves state-of-the-art accuracy on AIG property estimation tasks but also significantly improves training efficiency and scalability. These advantages stem directly from its targeted architectural innovations designed to address the structural heterogeneity and global information challenges inherent to AIG-based circuits.

\begin{table}[t]
  \centering
  \caption{The performance on the SPP Task}
  \label{tab:1}
\begin{tabular}{lcccc}
\toprule

Model          & Params($\times10^6 $) & Mem(GB) $\downarrow$& Avg.time(s) $\downarrow$& MAE $\downarrow$   \\
\midrule
GCN~\cite{kipf2016semi}            & 0.06   & 0.37    & 0.66        & 0.0623 \\
GraphSAGE~\cite{hamilton2017inductive}      & 0.09   & 0.40    & 0.89        & 0.0895 \\
\midrule
DeepGate~\cite{li2022deepgate}       & 0.07   & 0.31    & 56.11       & 0.1241 \\
DeepGate2~\cite{shi2023deepgate2}      & 1.34   & 0.89    & 47.61       & 0.0221 \\
DeepGate3~\cite{shi2024deepgate3}      & 8.56  & 63.46   & 13.79       & 0.0346 \\
HOGA~\cite{deng2024less}      & 1.18  & 3.85   & 5.27       & 0.1344 \\
PolarGate(L=3) & 0.41   & 0.54    & 3.52        & 0.0437 \\
PolarGate(L=9)~\cite{liu2024polargate} & 0.98   & 0.67    & 6.12        & 0.0097 \\
\midrule
FuncGNN  & 0.96   & \textbf{0.45}    & \textbf{3.02}        & \textbf{0.0095} \\

\bottomrule
\end{tabular}
\end{table}

\subsection{Answer to RQ2: Performance on the TTDP Task}
\quad
\textbf{Motivation.}The objective of this research question is to evaluate the performance of the proposed method on the TTDP task and compare it with state-of-the-art methods.

\textbf{Methodology.} Consistent with RQ1, the performance of the method on the TTDP task was evaluated using four metrics: MAE, Params, Mem, and Avg.time.

\textbf{Results.} 
Table~\ref{tab.ttdp} summarizes the performance of \modelname\xspace on the TTDP task. Across all evaluated metrics, \modelname\xspace consistently outperforms prior methods in both predictive accuracy and computational efficiency, demonstrating its effectiveness for functionality-aware AIG representation learning.

Compared to the DeepGate series, which rely heavily on attention mechanisms, \modelname\xspace achieves up to a 56.06\% reduction in MAE, while also exhibiting significantly faster training speeds. This performance gain is largely due to the Hybrid Feature Aggregation Component, which uses lightweight message-passing rather than attention, greatly reducing computational overhead while maintaining functional sensitivity.
Compared to PolarGate (L=9), which serves as a strong message-passing baseline with a comparable parameter count, \modelname\xspace achieves notable improvements. It reduces memory usage by 34.85\%, lowers average training time by 50.55\%, and enhances accuracy with an 18.71\% reduction in MAE. These gains are primarily attributed to the Global Context Normalization Component, which dynamically adjusts feature scaling based on logic-specific statistics. This mechanism enables better generalization across heterogeneous circuits while maintaining computational efficiency.
When compared with simpler graph models such as GCN and GraphSAGE, \modelname\xspace reduces MAE by 59.68\% and 44.08\% respectively. These results emphasize the effectiveness of the Multi-Layer Integration Component, which preserves and fuses multi-depth representations, mitigating over-squashing and enabling richer functional abstraction across circuit depths.

Together, these results affirm that \modelname\xspace not only delivers state-of-the-art predictive performance for the TTDP task but also scales efficiently to large and complex AIGs. It strikes an optimal balance between accuracy and resource consumption through its carefully crafted architectural components.

\begin{table}[t]
\centering
\caption{The performance on the TTDP Task}\label{tab.ttdp}
\begin{tabular}{lcccc}
\toprule
Model          & Params($\times10^6 $) & Mem(GB) $\downarrow$& Avg.time(s) $\downarrow$& MAE $\downarrow$   \\
\midrule
GCN~\cite{kipf2016semi}            & 0.06   & 0.39    & 0.79        & 0.4581 \\
GraphSAGE~\cite{hamilton2017inductive}      & 0.09   & 0.41    & 1.21        & 0.3303 \\
\midrule
DeepGate~\cite{li2022deepgate}       & 0.17   & 0.33    & 58.23       & 0.4203 \\
DeepGate2~\cite{shi2023deepgate2}      & 1.34   & 0.84    & 49.95       & 0.3912 \\
DeepGate3~\cite{shi2024deepgate3}      & 8.56  & 63.49   & 14.23       & 0.4367 \\
HOGA~\cite{deng2024less}      & 1.18  & 3.97   & 5.44       & 0.3244 \\
PolarGate(L=3) & 0.41   & 0.54    & 3.66        & 0.2513 \\
PolarGate(L=9)~\cite{liu2024polargate} & 0.98   & 0.66    & 6.39        & 0.2272 \\
\midrule
FuncGNN  & 0.96   & \textbf{0.43}    & \textbf{3.16}        &\textbf{0.1847} \\
\bottomrule
\end{tabular}
\end{table}

\subsection{Answer to RQ3: Performance in Stability}
\quad
\textbf{Motivation.}This research question investigates whether \modelname\xspace can maintain stable performance under the inherent structural variability of AIGs, especially when trained on a limited dataset. In real-world EDA scenarios, acquiring large-scale annotated data is often costly or impractical. Under such constraints, models with low robustness may exhibit significant performance fluctuations across different training sets. Therefore, we assess whether \modelname\xspace can mitigate training instability and consistently generalize across diverse circuit structures.

\textbf{Methodology.} 
To rigorously evaluate the robustness of each model under limited-data conditions, all hyperparameters and training configurations were fixed, with the random seed serving as the sole variable. Each random seed determines not only the data shuffling order but also the partitioning of training and test sets as well as the composition of mini-batches, thereby inducing diverse training subsets. Ten distinct seeds (42, 42×2, 42×3, ..., 42×10) were employed for training, with no early stopping applied. For PolarGate, L=9.

\begin{figure}[htbp]
  \centering
  \includegraphics[width=\textwidth]{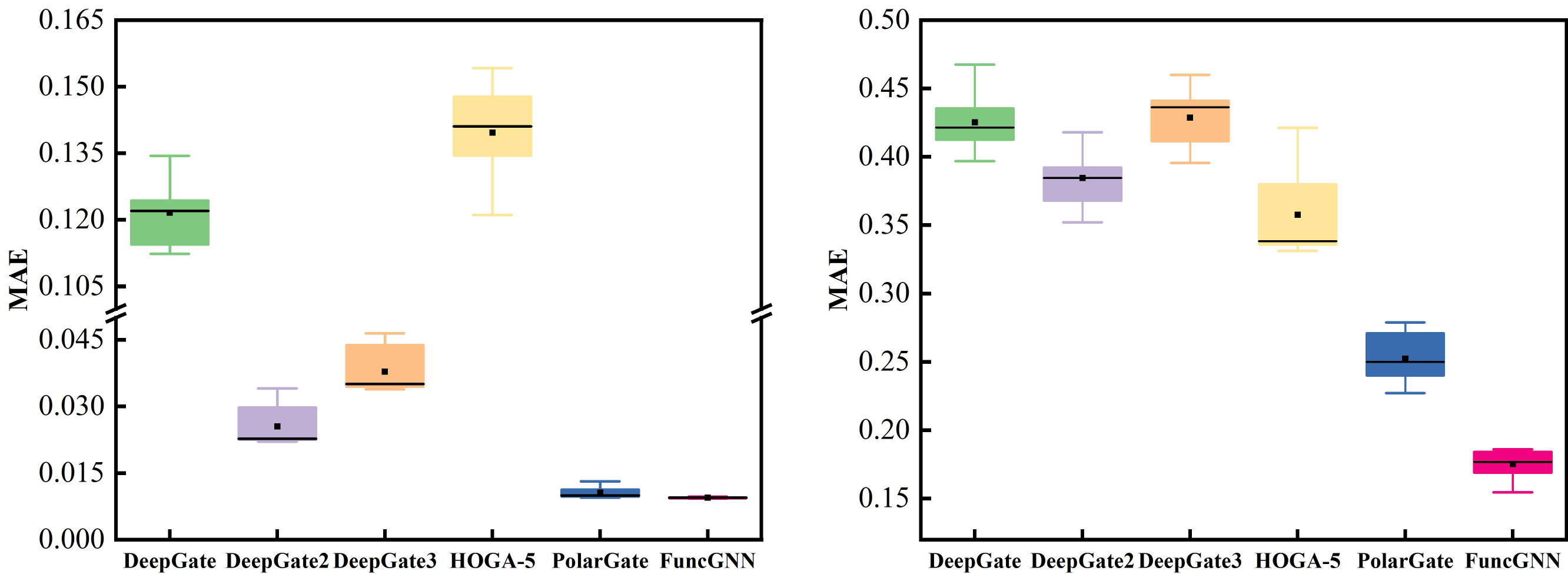}
  
  \begin{minipage}[t]{0.48\textwidth}
    \centering
    \textbf{(a)} SPP Task.
  \end{minipage}
  \hfill
  \begin{minipage}[t]{0.48\textwidth}
    \centering
    \textbf{(b)} TTDP Task.
  \end{minipage}
  \caption{Stability analysis on SPP and TTDP tasks.}
  \label{fig:rq3_both}
\end{figure}

This experimental design emulates realistic scenarios characterized by scarce and structurally diverse data. A model demonstrating strong robustness is expected to yield minimal variation in MAE across different seeds, whereas significant fluctuations would indicate limited adaptability to the structural heterogeneity of AIGs and a higher susceptibility to data shifts. Furthermore, the training setup was deliberately designed to be highly sparse, with only 496 AIGs allocated to training and 8941 AIGs to testing, resulting in an approximate train-to-test ratio of 1:18.

\textbf{Results.} 
\figurename~\ref{fig:rq3_both} reports the MAE variance of \modelname\xspace and state-of-the-art methods across ten different random seeds. On the SPP task, \modelname\xspace achieves the lowest MAE interquartile range (IQR) of 0.000175 and a whisker span of 0.0004, representing a dramatic reduction of 96.92\% in IQR and 97.14\% in span compared to the best DeepGate variant (IQR = 0.0057, span = 0.012), and a striking advantage over HOGA(IQR = 0.012, span = 0.0331) and PolarGate (IQR = 0.00155, span = 0.0036). These results demonstrate FuncGNN’s exceptional training stability and minimal sensitivity to initialization, particularly in the SPP task. On the TTDP task, FuncGNN also achieves a significant stability improvement, with an IQR of 0.0126 and a span of 0.0314. This marks a 35.38\% reduction in IQR and 51.63\% reduction in span compared to the best DeepGate result (IQR = 0.0195, span = 0.0643), and a noticeable advantage over HOGA(IQR = 0.0344, span = 0.0901) and PolarGate (IQR = 0.0268, span = 0.0516).

These improvements stem from the joint effect of Hybrid Feature Aggregation and Global Context Normalization, which together stabilize learning by preserving multi-scale structural signals and dynamically adapting to inter-graph logic variation.
\modelname\xspace consistently generalizes well under this setup, highlighting its robustness, high data efficiency, and ability to abstract functional patterns from limited supervision.

\begin{figure}[htbp]
  \centering
  \includegraphics[width=\textwidth]{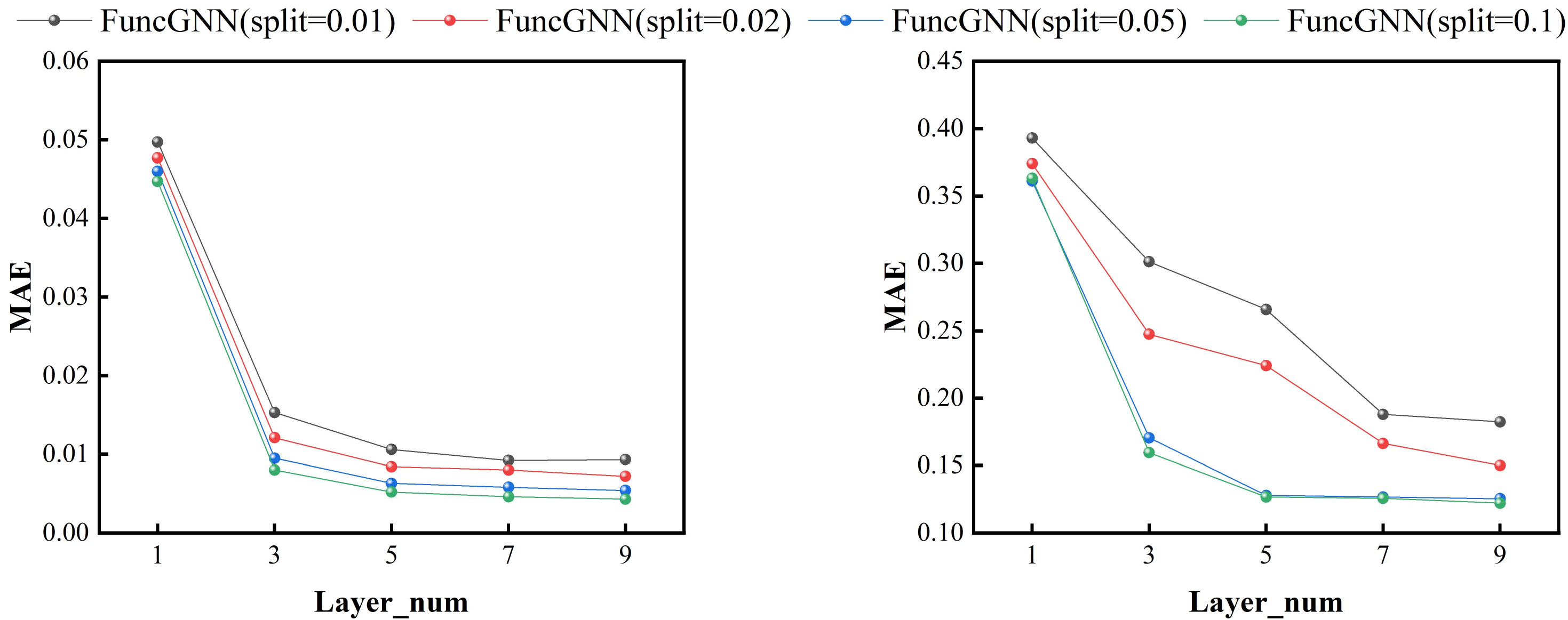}
  
  \begin{minipage}[t]{0.48\textwidth}
    \centering
    \textbf{(a)} SPP Task.
  \end{minipage}
  \hfill
  \begin{minipage}[t]{0.48\textwidth}
    \centering
    \textbf{(b)} TTDP Task.
  \end{minipage}
  \caption{Comparison of different layers on SPP and TTDP tasks.}
  \label{fig:rq4_both}
\end{figure}

\subsection{Answer to RQ4: Model Performance Across Different Layer Counts}
\quad
\textbf{Motivation.} GNNs inherently suffer from the over-squashing problem, where information from distant nodes is compressed into fixed-size embeddings as messages propagate through multiple layers. This compression limits the effective receptive field and leads to the loss of crucial global structural and functional information in large-scale circuit graphs such as AIGs. As a result, existing GNN-based methods struggle to accurately capture long-range dependencies critical for circuit property estimation. This research question aims to experimentally validate the ability of \modelname\xspace to mitigate the loss of global graph information due to over-squashing.
Attention-based methods are not included in this comparison, as they are generally more resilient to over-squashing and rely on fundamentally different global aggregation mechanisms.

\textbf{Methodology.} To assess the performance of the method under varying model complexities and data volumes, all hyperparameters except layer count were fixed, with model layer count and data splitting ratios as variables. Increasing the layer count typically elevates model complexity; however, deep models often suffer from over-squashing, leading to degraded accuracy despite higher complexity. To thoroughly evaluate whether \modelname\xspace overcomes over-squashing, experiments were conducted with odd layer counts (L=1, 3, 5, 7, 9) 
and multiple data splitting ratios (training and validation set proportions of 0.01, 0.02, 0.05, and 0.1). By monitoring MAE trends across these configurations, we evaluated \modelname’s\xspace robustness to over-squashing and its scalability with increasing depth.

\textbf{Results.} \figurename~\ref{fig:rq4_both} presents the performance of \modelname\xspace across different layer counts and data splits. As the number of layers increases, \modelname\xspace exhibits a steady decline in MAE for both the SPP and TTDP tasks, indicating consistent performance improvement. This trend is particularly evident under the 0.05 data split ratio, where MAE shows a noticeable reduction across different layer counts. Specifically, from L=1 to L=3, MAE decreases by approximately 79.3\% and 52.8\% for SPP and TTDP, respectively; from L=3 to L=5, it drops a further 33.7\% and 25.0\%; from L=5 to L=7, an additional 7.9\% and 0.9\%; and from L=7 to L=9, a total reduction of 6.9\% and 0.9\% is achieved. These results demonstrate that \modelname\xspace effectively leverages deeper architectures to integrate richer circuit features, thereby enhancing representational capacity and mitigating the challenge of over-squashing. Moreover, even under smaller data split ratios (such as 0.01 and 0.02), \modelname\xspace maintains stable and accurate performance, highlighting its robustness and sample efficiency in low-data regimes.

This consistency is primarily attributed to the proposed Multi-Layer Integration Component, which retains and adaptively fuses intermediate representations from all layers. By preserving rich hierarchical semantics across the network depth, this component enables \modelname\xspace to effectively aggregate both local and global information, ensuring high-quality circuit representations and strong generalization performance even under limited supervision.

\begin{table}[]
\centering
\caption{Performance On Ablation Study}\label{tab:abl_study}
\begin{tabular}{lll}
\toprule
Strategy                       & SPP    & TTDP   \\
\midrule
FuncGNN (w/o Component1 + GCN)  & 0.0178 & 0.3361 \\
FuncGNN (w/o Component2)        & 0.0113 & 0.2142 \\
FuncGNN (w/o Component3)        & 0.0231  & 0.3491 \\
FuncGNN (with simple GraphNorm) & 0.0107 & 0.2241 \\
FuncGNN (Full)                  & \textbf{0.0095} & \textbf{0.1847}\\
\bottomrule
\end{tabular}
\end{table}

\subsection{Answer to RQ5: Performance in Ablation Studies}
\quad
\textbf{Motivation.} This research question aims to validate the necessity of each component in \modelname\xspace and quantitatively assess their individual contributions to model performance.

\textbf{Methodology.} Ablation studies were conducted on \modelname\xspace to evaluate the impact of its key components. Individual components were either removed or replaced, and the resulting performance was assessed on the SPP and TTDP tasks using MAE as the primary metric. The experiments maintained consistent hyperparameters and a data splitting ratio of 0.05-0.05-0.9, as used in prior sections. 

\textbf{Results.} The results of the ablation studies are presented in Table \ref{tab:abl_study}, demonstrating that each component significantly influences \modelname’s performance. Specifically:
\begin{itemize}
    \item Replacing Component 1 with GCN led to MAE increases of 87.37\% and 81.97\% on the SPP and TTDP tasks, respectively, indicating severe performance degradation and underscoring the critical role of Component 1.
    \item Removing Component 2 resulted in MAE increases of 18.95\% and 15.97\% on the SPP and TTDP tasks, respectively, confirming its indispensability to model performance.
    \item Eliminating Component 3 caused MAE increases of 142.11\% and 89.01\% on the SPP and TTDP tasks, respectively, highlighting its importance in capturing global information.
    \item Substituting Component 2 with an unoptimized GraphNorm increased MAE by 12.63\% and 21.33\% on the SPP and TTDP tasks, respectively, demonstrating the performance benefits of incorporating global gate proportions into conditional normalization, particularly pronounced in the TTDP task.
\end{itemize}

In summary, the ablation studies confirm the essential role of each component in \modelname, collectively enhancing its robust and efficient AIG representation learning across diverse tasks.

\section{Conclusions and Future Work}
\label{sec:conclusion}

In this paper, we propose \modelname, a lightweight and robust framework for general circuit representation learning. It is designed to support a wide range of EDA tasks by effectively modeling AIG structures. \modelname\xspace combines three key components: Hybrid Feature Aggregation, Global Context Normalization, and Multi-Layer Integration. These components work together to improve learning efficiency, reduce computational costs, and enhance generalization, especially under conditions of structural diversity and limited training data.

Extensive experiments show that \modelname\xspace achieves superior performance compared to state-of-the-art baselines. It consistently delivers higher accuracy, better training stability, and greater scalability. This is particularly evident in scenarios with deep architectures or extreme train/test splits, where efficient and reliable circuit representation is critical for downstream EDA tasks such as logic synthesis and property prediction.

Future work includes validating \modelname\xspace on structurally diverse circuit datasets and broadening its application to a wider range of EDA tasks. Additionally, there will be attempts to theoretically demonstrate how incorporating AIG ratio information as a global prior may contribute to constructing globally stable feature representations.

\begin{acks}

This work was supported by the National Natural Science Foundation of China (No.62472062, No.62202079), the Dalian Science and technology Innovation Fund project (No.2024JJ12GX022), the key research and development project of Liaoning Province (No.2024JH2/102400059).

\end{acks}
\bibliographystyle{ACM-Reference-Format}
\bibliography{1-TRETS/GUIDENCE}


\begin{thebibliography}{34}


\ifx \showCODEN    \undefined \def \showCODEN     #1{\unskip}     \fi
\ifx \showDOI      \undefined \def \showDOI       #1{#1}\fi
\ifx \showISBNx    \undefined \def \showISBNx     #1{\unskip}     \fi
\ifx \showISBNxiii \undefined \def \showISBNxiii  #1{\unskip}     \fi
\ifx \showISSN     \undefined \def \showISSN      #1{\unskip}     \fi
\ifx \showLCCN     \undefined \def \showLCCN      #1{\unskip}     \fi
\ifx \shownote     \undefined \def \shownote      #1{#1}          \fi
\ifx \showarticletitle \undefined \def \showarticletitle #1{#1}   \fi
\ifx \showURL      \undefined \def \showURL       {\relax}        \fi
\providecommand\bibfield[2]{#2}
\providecommand\bibinfo[2]{#2}
\providecommand\natexlab[1]{#1}
\providecommand\showeprint[2][]{arXiv:#2}

\bibitem[Albrecht(2005)]%
        {albrecht2005iwls}
\bibfield{author}{\bibinfo{person}{Christoph Albrecht}.} \bibinfo{year}{2005}\natexlab{}.
\newblock \showarticletitle{IWLS 2005 benchmarks}. In \bibinfo{booktitle}{\emph{International Workshop for Logic Synthesis (IWLS)}}, Vol.~\bibinfo{volume}{9}.
\newblock


\bibitem[Alon and Yahav(2020)]%
        {alon2020bottleneck}
\bibfield{author}{\bibinfo{person}{Uri Alon} {and} \bibinfo{person}{Eran Yahav}.} \bibinfo{year}{2020}\natexlab{}.
\newblock \showarticletitle{On the bottleneck of graph neural networks and its practical implications}.
\newblock \bibinfo{journal}{\emph{arXiv preprint arXiv:2006.05205}} (\bibinfo{year}{2020}).
\newblock


\bibitem[Amar{\'u} et~al\mbox{.}(2015)]%
        {amaru2015epfl}
\bibfield{author}{\bibinfo{person}{Luca Amar{\'u}}, \bibinfo{person}{Pierre-Emmanuel Gaillardon}, {and} \bibinfo{person}{Giovanni De~Micheli}.} \bibinfo{year}{2015}\natexlab{}.
\newblock \showarticletitle{The EPFL combinational benchmark suite}. In \bibinfo{booktitle}{\emph{Proceedings of the 24th International Workshop on Logic \& Synthesis (IWLS)}}.
\newblock


\bibitem[Bevilacqua et~al\mbox{.}(2021)]%
        {bevilacqua2021size}
\bibfield{author}{\bibinfo{person}{Beatrice Bevilacqua}, \bibinfo{person}{Yangze Zhou}, {and} \bibinfo{person}{Bruno Ribeiro}.} \bibinfo{year}{2021}\natexlab{}.
\newblock \showarticletitle{Size-invariant graph representations for graph classification extrapolations}. In \bibinfo{booktitle}{\emph{International Conference on Machine Learning}}. PMLR, \bibinfo{pages}{837--851}.
\newblock


\bibitem[Brayton and Mishchenko(2010)]%
        {brayton2010abc}
\bibfield{author}{\bibinfo{person}{Robert Brayton} {and} \bibinfo{person}{Alan Mishchenko}.} \bibinfo{year}{2010}\natexlab{}.
\newblock \showarticletitle{ABC: An academic industrial-strength verification tool}. In \bibinfo{booktitle}{\emph{Computer Aided Verification: 22nd International Conference, CAV 2010, Edinburgh, UK, July 15-19, 2010. Proceedings 22}}. Springer, \bibinfo{pages}{24--40}.
\newblock


\bibitem[Cai et~al\mbox{.}(2021)]%
        {cai2021graphnorm}
\bibfield{author}{\bibinfo{person}{Tianle Cai}, \bibinfo{person}{Shengjie Luo}, \bibinfo{person}{Keyulu Xu}, \bibinfo{person}{Di He}, \bibinfo{person}{Tie-yan Liu}, {and} \bibinfo{person}{Liwei Wang}.} \bibinfo{year}{2021}\natexlab{}.
\newblock \showarticletitle{Graphnorm: A principled approach to accelerating graph neural network training}. In \bibinfo{booktitle}{\emph{International Conference on Machine Learning}}. PMLR, \bibinfo{pages}{1204--1215}.
\newblock


\bibitem[Davidson(1999)]%
        {davidson1999characteristics}
\bibfield{author}{\bibinfo{person}{Scott Davidson}.} \bibinfo{year}{1999}\natexlab{}.
\newblock \showarticletitle{Characteristics of the ITC’99 benchmark circuits}. In \bibinfo{booktitle}{\emph{IEEE International Test Synthesis Workshop (ITSW)}}. \bibinfo{pages}{87}.
\newblock


\bibitem[De~Vries et~al\mbox{.}(2017)]%
        {de2017modulating}
\bibfield{author}{\bibinfo{person}{Harm De~Vries}, \bibinfo{person}{Florian Strub}, \bibinfo{person}{J{\'e}r{\'e}mie Mary}, \bibinfo{person}{Hugo Larochelle}, \bibinfo{person}{Olivier Pietquin}, {and} \bibinfo{person}{Aaron~C Courville}.} \bibinfo{year}{2017}\natexlab{}.
\newblock \showarticletitle{Modulating early visual processing by language}.
\newblock \bibinfo{journal}{\emph{Advances in neural information processing systems}}  \bibinfo{volume}{30} (\bibinfo{year}{2017}).
\newblock


\bibitem[Deng et~al\mbox{.}(2024)]%
        {deng2024less}
\bibfield{author}{\bibinfo{person}{Chenhui Deng}, \bibinfo{person}{Zichao Yue}, \bibinfo{person}{Cunxi Yu}, \bibinfo{person}{Gokce Sarar}, \bibinfo{person}{Ryan Carey}, \bibinfo{person}{Rajeev Jain}, {and} \bibinfo{person}{Zhiru Zhang}.} \bibinfo{year}{2024}\natexlab{}.
\newblock \showarticletitle{Less is more: Hop-wise graph attention for scalable and generalizable learning on circuits}. In \bibinfo{booktitle}{\emph{Proceedings of the 61st ACM/IEEE Design Automation Conference}}. \bibinfo{pages}{1--6}.
\newblock


\bibitem[Fang et~al\mbox{.}(2025)]%
        {fang2025survey}
\bibfield{author}{\bibinfo{person}{Wenji Fang}, \bibinfo{person}{Jing Wang}, \bibinfo{person}{Yao Lu}, \bibinfo{person}{Shang Liu}, \bibinfo{person}{Yuchao Wu}, \bibinfo{person}{Yuzhe Ma}, {and} \bibinfo{person}{Zhiyao Xie}.} \bibinfo{year}{2025}\natexlab{}.
\newblock \showarticletitle{A survey of circuit foundation model: Foundation ai models for vlsi circuit design and eda}.
\newblock \bibinfo{journal}{\emph{arXiv preprint arXiv:2504.03711}} (\bibinfo{year}{2025}).
\newblock


\bibitem[FuncGNN(2025)]%
        {funcgnn}
\bibfield{author}{\bibinfo{person}{FuncGNN}.} \bibinfo{year}{2025}\natexlab{}.
\newblock \bibinfo{booktitle}{\emph{FuncGNN Project Repository}}.
\newblock
\urldef\tempurl%
\url{https://github.com/Vandbs/FuncGNN}
\showURL{%
\tempurl}


\bibitem[Hamilton et~al\mbox{.}(2017)]%
        {hamilton2017inductive}
\bibfield{author}{\bibinfo{person}{Will Hamilton}, \bibinfo{person}{Zhitao Ying}, {and} \bibinfo{person}{Jure Leskovec}.} \bibinfo{year}{2017}\natexlab{}.
\newblock \showarticletitle{Inductive representation learning on large graphs}.
\newblock \bibinfo{journal}{\emph{Advances in neural information processing systems}}  \bibinfo{volume}{30} (\bibinfo{year}{2017}).
\newblock


\bibitem[Hendrycks and Gimpel(2016)]%
        {hendrycks2016gaussian}
\bibfield{author}{\bibinfo{person}{Dan Hendrycks} {and} \bibinfo{person}{Kevin Gimpel}.} \bibinfo{year}{2016}\natexlab{}.
\newblock \showarticletitle{Gaussian error linear units (gelus)}.
\newblock \bibinfo{journal}{\emph{arXiv preprint arXiv:1606.08415}} (\bibinfo{year}{2016}).
\newblock


\bibitem[Huang et~al\mbox{.}(2021)]%
        {huang2021machine}
\bibfield{author}{\bibinfo{person}{Guyue Huang}, \bibinfo{person}{Jingbo Hu}, \bibinfo{person}{Yifan He}, \bibinfo{person}{Jialong Liu}, \bibinfo{person}{Mingyuan Ma}, \bibinfo{person}{Zhaoyang Shen}, \bibinfo{person}{Juejian Wu}, \bibinfo{person}{Yuanfan Xu}, \bibinfo{person}{Hengrui Zhang}, \bibinfo{person}{Kai Zhong}, {et~al\mbox{.}}} \bibinfo{year}{2021}\natexlab{}.
\newblock \showarticletitle{Machine learning for electronic design automation: A survey}.
\newblock \bibinfo{journal}{\emph{ACM Transactions on Design Automation of Electronic Systems (TODAES)}} \bibinfo{volume}{26}, \bibinfo{number}{5} (\bibinfo{year}{2021}), \bibinfo{pages}{1--46}.
\newblock


\bibitem[Joshi et~al\mbox{.}(2022)]%
        {joshi2022learning}
\bibfield{author}{\bibinfo{person}{Chaitanya~K Joshi}, \bibinfo{person}{Quentin Cappart}, \bibinfo{person}{Louis-Martin Rousseau}, {and} \bibinfo{person}{Thomas Laurent}.} \bibinfo{year}{2022}\natexlab{}.
\newblock \showarticletitle{Learning the travelling salesperson problem requires rethinking generalization}.
\newblock \bibinfo{journal}{\emph{Constraints}} \bibinfo{volume}{27}, \bibinfo{number}{1} (\bibinfo{year}{2022}), \bibinfo{pages}{70--98}.
\newblock


\bibitem[Kipf and Welling(2016)]%
        {kipf2016semi}
\bibfield{author}{\bibinfo{person}{Thomas~N Kipf} {and} \bibinfo{person}{Max Welling}.} \bibinfo{year}{2016}\natexlab{}.
\newblock \showarticletitle{Semi-supervised classification with graph convolutional networks}.
\newblock \bibinfo{journal}{\emph{arXiv preprint arXiv:1609.02907}} (\bibinfo{year}{2016}).
\newblock


\bibitem[Li et~al\mbox{.}(2022)]%
        {li2022deepgate}
\bibfield{author}{\bibinfo{person}{Min Li}, \bibinfo{person}{Sadaf Khan}, \bibinfo{person}{Zhengyuan Shi}, \bibinfo{person}{Naixing Wang}, \bibinfo{person}{Huang Yu}, {and} \bibinfo{person}{Qiang Xu}.} \bibinfo{year}{2022}\natexlab{}.
\newblock \showarticletitle{Deepgate: Learning neural representations of logic gates}. In \bibinfo{booktitle}{\emph{Proceedings of the 59th ACM/IEEE Design Automation Conference}}. \bibinfo{pages}{667--672}.
\newblock


\bibitem[Liu et~al\mbox{.}(2024)]%
        {liu2024polargate}
\bibfield{author}{\bibinfo{person}{Jiawei Liu}, \bibinfo{person}{Jianwang Zhai}, \bibinfo{person}{Mingyu Zhao}, \bibinfo{person}{Zhe Lin}, \bibinfo{person}{Bei Yu}, {and} \bibinfo{person}{Chuan Shi}.} \bibinfo{year}{2024}\natexlab{}.
\newblock \showarticletitle{Polargate: Breaking the functionality representation bottleneck of and-inverter graph neural network}. In \bibinfo{booktitle}{\emph{Proceedings of the 43rd IEEE/ACM International Conference on Computer-Aided Design}}. \bibinfo{pages}{1--9}.
\newblock


\bibitem[Mishchenko et~al\mbox{.}(2006)]%
        {mishchenko2006dag}
\bibfield{author}{\bibinfo{person}{Alan Mishchenko}, \bibinfo{person}{Satrajit Chatterjee}, {and} \bibinfo{person}{Robert Brayton}.} \bibinfo{year}{2006}\natexlab{}.
\newblock \showarticletitle{DAG-aware AIG rewriting a fresh look at combinational logic synthesis}. In \bibinfo{booktitle}{\emph{Proceedings of the 43rd annual Design Automation Conference}}. \bibinfo{pages}{532--535}.
\newblock


\bibitem[Mishchenko et~al\mbox{.}(2005)]%
        {mishchenko2005fraigs}
\bibfield{author}{\bibinfo{person}{Alan Mishchenko}, \bibinfo{person}{Satrajit Chatterjee}, \bibinfo{person}{Roland Jiang}, {and} \bibinfo{person}{Robert~K Brayton}.} \bibinfo{year}{2005}\natexlab{}.
\newblock \bibinfo{booktitle}{\emph{FRAIGs: A unifying representation for logic synthesis and verification}}.
\newblock \bibinfo{type}{{T}echnical {R}eport}. \bibinfo{institution}{ERL Technical Report}.
\newblock


\bibitem[Perez et~al\mbox{.}(2018)]%
        {perez2018film}
\bibfield{author}{\bibinfo{person}{Ethan Perez}, \bibinfo{person}{Florian Strub}, \bibinfo{person}{Harm De~Vries}, \bibinfo{person}{Vincent Dumoulin}, {and} \bibinfo{person}{Aaron Courville}.} \bibinfo{year}{2018}\natexlab{}.
\newblock \showarticletitle{Film: Visual reasoning with a general conditioning layer}. In \bibinfo{booktitle}{\emph{Proceedings of the AAAI conference on artificial intelligence}}, Vol.~\bibinfo{volume}{32}.
\newblock


\bibitem[Rapp et~al\mbox{.}(2021)]%
        {rapp2021mlcad}
\bibfield{author}{\bibinfo{person}{Martin Rapp}, \bibinfo{person}{Hussam Amrouch}, \bibinfo{person}{Yibo Lin}, \bibinfo{person}{Bei Yu}, \bibinfo{person}{David~Z Pan}, \bibinfo{person}{Marilyn Wolf}, {and} \bibinfo{person}{J{\"o}rg Henkel}.} \bibinfo{year}{2021}\natexlab{}.
\newblock \showarticletitle{MLCAD: A survey of research in machine learning for CAD keynote paper}.
\newblock \bibinfo{journal}{\emph{IEEE Transactions on Computer-Aided Design of Integrated Circuits and Systems}} \bibinfo{volume}{41}, \bibinfo{number}{10} (\bibinfo{year}{2021}), \bibinfo{pages}{3162--3181}.
\newblock


\bibitem[Scarselli et~al\mbox{.}(2008)]%
        {scarselli2008graph}
\bibfield{author}{\bibinfo{person}{Franco Scarselli}, \bibinfo{person}{Marco Gori}, \bibinfo{person}{Ah~Chung Tsoi}, \bibinfo{person}{Markus Hagenbuchner}, {and} \bibinfo{person}{Gabriele Monfardini}.} \bibinfo{year}{2008}\natexlab{}.
\newblock \showarticletitle{The graph neural network model}.
\newblock \bibinfo{journal}{\emph{IEEE transactions on neural networks}} \bibinfo{volume}{20}, \bibinfo{number}{1} (\bibinfo{year}{2008}), \bibinfo{pages}{61--80}.
\newblock


\bibitem[Shi et~al\mbox{.}(2023)]%
        {shi2023deepgate2}
\bibfield{author}{\bibinfo{person}{Zhengyuan Shi}, \bibinfo{person}{Hongyang Pan}, \bibinfo{person}{Sadaf Khan}, \bibinfo{person}{Min Li}, \bibinfo{person}{Yi Liu}, \bibinfo{person}{Junhua Huang}, \bibinfo{person}{Hui-Ling Zhen}, \bibinfo{person}{Mingxuan Yuan}, \bibinfo{person}{Zhufei Chu}, {and} \bibinfo{person}{Qiang Xu}.} \bibinfo{year}{2023}\natexlab{}.
\newblock \showarticletitle{Deepgate2: Functionality-aware circuit representation learning}. In \bibinfo{booktitle}{\emph{2023 IEEE/ACM International Conference on Computer Aided Design (ICCAD)}}. IEEE, \bibinfo{pages}{1--9}.
\newblock


\bibitem[Shi et~al\mbox{.}(2024)]%
        {shi2024deepgate3}
\bibfield{author}{\bibinfo{person}{Zhengyuan Shi}, \bibinfo{person}{Ziyang Zheng}, \bibinfo{person}{Sadaf Khan}, \bibinfo{person}{Jianyuan Zhong}, \bibinfo{person}{Min Li}, {and} \bibinfo{person}{Qiang Xu}.} \bibinfo{year}{2024}\natexlab{}.
\newblock \showarticletitle{Deepgate3: Towards scalable circuit representation learning}. In \bibinfo{booktitle}{\emph{Proceedings of the 43rd IEEE/ACM International Conference on Computer-Aided Design}}. \bibinfo{pages}{1--9}.
\newblock


\bibitem[Srivastava et~al\mbox{.}(2014)]%
        {srivastava2014dropout}
\bibfield{author}{\bibinfo{person}{Nitish Srivastava}, \bibinfo{person}{Geoffrey Hinton}, \bibinfo{person}{Alex Krizhevsky}, \bibinfo{person}{Ilya Sutskever}, {and} \bibinfo{person}{Ruslan Salakhutdinov}.} \bibinfo{year}{2014}\natexlab{}.
\newblock \showarticletitle{Dropout: a simple way to prevent neural networks from overfitting}.
\newblock \bibinfo{journal}{\emph{The journal of machine learning research}} \bibinfo{volume}{15}, \bibinfo{number}{1} (\bibinfo{year}{2014}), \bibinfo{pages}{1929--1958}.
\newblock


\bibitem[Subramanyan et~al\mbox{.}(2013)]%
        {subramanyan2013reverse}
\bibfield{author}{\bibinfo{person}{Pramod Subramanyan}, \bibinfo{person}{Nestan Tsiskaridze}, \bibinfo{person}{Wenchao Li}, \bibinfo{person}{Adria Gasc{\'o}n}, \bibinfo{person}{Wei~Yang Tan}, \bibinfo{person}{Ashish Tiwari}, \bibinfo{person}{Natarajan Shankar}, \bibinfo{person}{Sanjit~A Seshia}, {and} \bibinfo{person}{Sharad Malik}.} \bibinfo{year}{2013}\natexlab{}.
\newblock \showarticletitle{Reverse engineering digital circuits using structural and functional analyses}.
\newblock \bibinfo{journal}{\emph{IEEE Transactions on Emerging Topics in Computing}} \bibinfo{volume}{2}, \bibinfo{number}{1} (\bibinfo{year}{2013}), \bibinfo{pages}{63--80}.
\newblock


\bibitem[Team(2025)]%
        {opencores}
\bibfield{author}{\bibinfo{person}{O. Team}.} \bibinfo{year}{Accessed: 2025}\natexlab{}.
\newblock \bibinfo{title}{Opencores}.
\newblock \bibinfo{howpublished}{\url{https://opencores.org/}}.
\newblock
\newblock
\shownote{Accessed: May 2025}.


\bibitem[Wu et~al\mbox{.}(2023)]%
        {wu2023gamora}
\bibfield{author}{\bibinfo{person}{Nan Wu}, \bibinfo{person}{Yingjie Li}, \bibinfo{person}{Cong Hao}, \bibinfo{person}{Steve Dai}, \bibinfo{person}{Cunxi Yu}, {and} \bibinfo{person}{Yuan Xie}.} \bibinfo{year}{2023}\natexlab{}.
\newblock \showarticletitle{Gamora: Graph learning based symbolic reasoning for large-scale boolean networks}. In \bibinfo{booktitle}{\emph{2023 60th ACM/IEEE Design Automation Conference (DAC)}}. IEEE, \bibinfo{pages}{1--6}.
\newblock


\bibitem[Xu et~al\mbox{.}(2018)]%
        {xu2018powerful}
\bibfield{author}{\bibinfo{person}{Keyulu Xu}, \bibinfo{person}{Weihua Hu}, \bibinfo{person}{Jure Leskovec}, {and} \bibinfo{person}{Stefanie Jegelka}.} \bibinfo{year}{2018}\natexlab{}.
\newblock \showarticletitle{How powerful are graph neural networks?}
\newblock \bibinfo{journal}{\emph{arXiv preprint arXiv:1810.00826}} (\bibinfo{year}{2018}).
\newblock


\bibitem[Xu et~al\mbox{.}(2025a)]%
        {xu2025simtam}
\bibfield{author}{\bibinfo{person}{Zhihao Xu}, \bibinfo{person}{Shikai Guo}, \bibinfo{person}{Xiaochen Li}, \bibinfo{person}{Zun Wang}, {and} \bibinfo{person}{He Jiang}.} \bibinfo{year}{2025}\natexlab{a}.
\newblock \showarticletitle{SIMTAM: Generation Diversity Test Programs for FPGA Simulation Tools Testing Via Timing Area Mutation}.
\newblock \bibinfo{journal}{\emph{ACM Transactions on Design Automation of Electronic Systems}} \bibinfo{volume}{30}, \bibinfo{number}{2} (\bibinfo{year}{2025}), \bibinfo{pages}{1--25}.
\newblock


\bibitem[Xu et~al\mbox{.}(2025b)]%
        {xu2025novel}
\bibfield{author}{\bibinfo{person}{Zhihao Xu}, \bibinfo{person}{Shikai Guo}, \bibinfo{person}{Guilin Zhao}, \bibinfo{person}{Peiyu Zou}, \bibinfo{person}{Xiaochen Li}, {and} \bibinfo{person}{He Jiang}.} \bibinfo{year}{2025}\natexlab{b}.
\newblock \showarticletitle{A novel HDL code generator for effectively testing FPGA logic synthesis compilers}.
\newblock \bibinfo{journal}{\emph{IEEE Transactions on Computer-Aided Design of Integrated Circuits and Systems}} (\bibinfo{year}{2025}).
\newblock


\bibitem[Yehudai et~al\mbox{.}(2021)]%
        {yehudai2021local}
\bibfield{author}{\bibinfo{person}{Gilad Yehudai}, \bibinfo{person}{Ethan Fetaya}, \bibinfo{person}{Eli Meirom}, \bibinfo{person}{Gal Chechik}, {and} \bibinfo{person}{Haggai Maron}.} \bibinfo{year}{2021}\natexlab{}.
\newblock \showarticletitle{From local structures to size generalization in graph neural networks}. In \bibinfo{booktitle}{\emph{International Conference on Machine Learning}}. PMLR, \bibinfo{pages}{11975--11986}.
\newblock


\bibitem[Zhang et~al\mbox{.}(2021)]%
        {zhang2021circuit}
\bibfield{author}{\bibinfo{person}{He-Teng Zhang}, \bibinfo{person}{Jie-Hong~R Jiang}, {and} \bibinfo{person}{Alan Mishchenko}.} \bibinfo{year}{2021}\natexlab{}.
\newblock \showarticletitle{A circuit-based SAT solver for logic synthesis}. In \bibinfo{booktitle}{\emph{2021 IEEE/ACM International Conference On Computer Aided Design (ICCAD)}}. IEEE, \bibinfo{pages}{1--6}.
\newblock


\end{thebibliography}

\end{document}